\documentclass[10pt, final, conference]{IEEEtran}

\IEEEoverridecommandlockouts
\usepackage{cite}
\usepackage{amsmath,amssymb,amsfonts}
\usepackage{graphicx}
\usepackage{textcomp}
\usepackage{xcolor}
\usepackage{subfig}
\usepackage{algorithm}
\usepackage{colortbl}
\usepackage{algpseudocode}
\usepackage{listings}
\usepackage{textcomp}
\usepackage{wrapfig}

\usepackage[letterpaper]{geometry}
\usepackage{hyperref}

\usepackage{tabularx}
\usepackage[figuresright]{rotating}
\RequirePackage{float}
\RequirePackage{caption}
\RequirePackage{pdflscape}
\usepackage[subfigure]{tocloft}
\usepackage{booktabs}
\usepackage{rotating}
\usepackage{adjustbox}

\definecolor{codegreen}{rgb}{0,0.6,0}
\definecolor{codegray}{rgb}{0.5,0.5,0.5}
\definecolor{codepurple}{rgb}{0.58,0,0.82}
\definecolor{backcolour}{rgb}{0.95,0.95,0.92}

\lstdefinestyle{mystyle}{
    backgroundcolor=\color{backcolour},   
    commentstyle=\color{codegreen},
    keywordstyle=\color{magenta},
    numberstyle=\tiny\color{codegray},
    stringstyle=\color{codepurple},
    basicstyle=\ttfamily\footnotesize,
    breakatwhitespace=false,         
    breaklines=true,                 
    captionpos=b,                    
    keepspaces=true,                 
    numbers=left,                    
    numbersep=5pt,                  
    showspaces=false,                
    showstringspaces=false,
    showtabs=false,                  
    tabsize=2
}

\def\BibTeX{{\rm B\kern-.05em{\sc i\kern-.025em b}\kern-.08em
    T\kern-.1667em\lower.7ex\hbox{E}\kern-.125emX}}
    
\definecolor{abstractbg}{rgb}{0.5000,0.74510,0.5000}
\definecolor{blu}{rgb}{0.8000,0.9900,0.8}

\begin{document}
\pagenumbering{roman}
\title{\textit{ChemTime}: Rapid and Early Classification for Multivariate Time Series Classification of Chemical Sensors

}

\makeatletter
\newcommand{\linebreakand}{%
  \end{@IEEEauthorhalign}
  \hfill\mbox{}\par
  \mbox{}\hfill\begin{@IEEEauthorhalign}
}
\makeatother

\author{\IEEEauthorblockN{Alexander M. Moore}
\IEEEauthorblockA{\textit{Worcester Polytechnic Institute} \\
Worcester MA, USA \\
ammoore@wpi.edu}
\and
\IEEEauthorblockN{Randy C. Paffenroth}
\IEEEauthorblockA{\textit{Worcester Polytechnic Institute} \\
Worcester MA, USA \\
rcpaffenroth@wpi.edu}
\linebreakand
\IEEEauthorblockN{Ken T. Ngo}
\IEEEauthorblockA{\textit{Chemical/Biological Innovative} \\ \textit{Material and Ensemble Development Team} \\
\textit{U.S. Army DEVCOM Soldier Center}\\
Natick MA, USA \\
ken.a.ngo.civ@army.mil}
\and
\IEEEauthorblockN{Joshua R. Uzarski}
\IEEEauthorblockA{\textit{Chemical/Biological Innovative} \\ \textit{Material and Ensemble Development Team} \\
\textit{U.S. Army DEVCOM Soldier Center}\\
Natick MA, USA \\
joshua.r.uzarski.civ@army.mil}

}

\makeatletter
\def\ps@IEEEtitlepagestyle{%
  \def\@oddfoot{\mycopyrightnotice}%
  \def\@evenfoot{}%
}
\def\mycopyrightnotice{%
  {\hfill \footnotesize 978-1-4673-9563-2/15/\$31.00 \copyright 2022 IEEE\hfill}
}
\makeatother

\maketitle

\begin{abstract}
Multivariate time series data are ubiquitous in the application of machine learning to problems in the physical sciences. Chemiresistive sensor arrays are highly promising in chemical detection tasks relevant to industrial, safety, and military applications. Sensor arrays are an inherently multivariate time series data collection tool which demand rapid and accurate classification of arbitrary chemical analytes. Previous research has benchmarked data-agnostic multivariate time series classifiers across diverse multivariate time series supervised tasks in order to find general-purpose classification algorithms. To our knowledge, there has yet to be an effort to survey machine learning and time series classification approaches to chemiresistive hardware sensor arrays for the detection of chemical analytes. In addition to benchmarking existing approaches to multivariate time series classifiers, we incorporate findings from a model survey to propose the novel \textit{ChemTime} approach to sensor array classification for chemical sensing.  We design experiments addressing the unique challenges of hardware sensor arrays classification including the rapid classification ability of classifiers and minimization of inference time while maintaining performance for deployed lightweight hardware sensing devices. We find that \textit{ChemTime} is uniquely positioned for the chemical sensing task by combining rapid and early classification of time series with beneficial inference and high accuracy.
\end{abstract}

\begin{IEEEkeywords}
Chemical sensors, machine learning, multivariate time series, representation learning
\end{IEEEkeywords}

\section{Introduction}\label{sec:introduction}

Chemical sensors use an array of chemiresistive sensors to produce characteristic resistance signals. Multivariate time series classifiers may be trained to classify the presence of chemical analytes given training examples containing characteristic responses. Chemical analyte detection is vital to industrial and military applications, and carries unique challenges for the application of multivariate time series classifiers. In order to approach the topic of optimal chemical discrimination for a novel hardware sensor array, we propose a variety of supervised learning experiments with additional modifications to account for the challenges unique to the chemical sensing space.

Our findings on the optimal discrimination of chemical analytes leads us to propose a new multivariate time series classifier designed explicitly for chemical sensing. \textit{ChemTime} utilizes inductive biases in the data and known chemical structures to better encode time series signals to a meaningful chemistry-informed latent space (Section \ref{sec:intro_chemtime}). We demonstrate that these changes yield a performant model which is significantly faster and more lightweight than comparable models while maintaining a high degree of accuracy (Section \ref{sec:bo_results}).

In order to mitigate the effects of model inductive bias, as well as inductive biases of sets of sensors, we consider an empirical survey of diverse models from the established and modern literature for multivariate time series classification as well as diverse sets of chemiresistive sensor coatings. We benchmark each model on a broad set of chemiresistive sensor arrays and investigate patterns in successful models across multiple hardware designs. In addition we supply experiments and results for a variety of specific tasks relevant to chemical sensing with chemiresistive sensor arrays, including limit of detection studies, analysis of the rapid classification abilities of models, and analysis on the time to train and infer on our suite of models.

\subsection{Contributions}\label{sec:contributions}
Our contributions to the application of machine learning to the applied sciences and chemical sensor array classification include the following:
\begin{enumerate}
    \item A benchmarking of modern competitive multivariate time series classifiers from the literature on eleven real-world chemiresistive sensor datasets for the discrimination of a particular chemical analyte (Section \ref{sec:bo_results}).

    \item A novel approach to the classification of multivariate time series for chemiresistive sensors based on an adaptation of transfer learning to molecular representation which improves the efficiency frontier for chemical sensing (Section \ref{sec:intro_chemtime}).

    \item Analysis for tasks dependent on the rapid classification of chemical analytes including an efficiency frontier for inference time vs. accuracy across the span of benchmarked models
\end{enumerate}

\subsection{Chemical Sensing}\label{sec:chemical_sensing}

Chemical sensors measure physical and chemical properties of analytes into measurable signals. Examples of chemical sensors include breathalyzers, carbon monoxide sensors, and electrochemical gas sensors \cite{rana2021modern}. The detection of particular chemical analytes is highly relevant in civilian  safety, manufacturing, and military applications \cite{wiederoder2017graphene, weiss2018applications, moore2023chemvise}. Chemiresistive sensor devices respond to chemical analytes by reporting changes in resistance through a coated resistor. Chemical analytes interact with sensors at the molecular level by bonding with the sensor coating, called adsorption. Analyte adsorption to the sensor coating causes the resistance through the sensing element to change as a function of the binding affinity between the analyte and surface. The binding affinity between analytes and coatings is affected by their molecular and polymer chemical properties which yields the characteristic response curve for the analyte.

Non-chemiresistive chemical sensors include ``infrared and Raman spectroscopy, ion mobility spectrometry, surface acoustic wave sensors,`` and other on-site testing technologies requiring complex instrumentation, high cost, and expert operating personnel \cite{wiederoder2017graphene}. These expenses and complications limit the utility of a deployed tool, and increase the cost and time requirements of gathering experimental data for hardware development and machine learning training. More portable sensors may be limited by responsiveness to select analytes and perform may perform poorly in the presence of obscurant chemicals \cite{pacsial2013chemical, wiederoder2017graphene, moore2022acgans}. Contemporary sensor arrays including those used to collect the experimental data discussed here address the challenges of rapid discrimination of multiple target analytes with low-cost, low-power, miniaturized sensors capable of analyte detection despite obscurants with an array of semi-selective chemical sensors that respond to many analytes simultaneously \cite{wiederoder2017graphene}.

We propose analyses of machine and deep learning classifiers trained and tested on real-world chemical sensor resistance data from 8-sensor chemiresistive arrays with chemically diverse coatings to maximize analyte discriminability as in \cite{nallon2016discrimination, wiederoder2017graphene, weiss2018applications}.

The unique coatings on the chemiresistive sensor cause lead to characteristic resistances of analytes which facilitate discrimination of the gas exposures. Figure \ref{fig:two_singles} shows an example of the contrast in sensor responses to $17.5\%$ Analyte A and $17.5\%$ Analyte B vapors given the same set of eight sensors with bespoke chemiresistive coatings. From the 8-channel characteristic signal machine learning models learn decision patterns for generalization to unseen testing samples.

\begin{figure*}
    \centering
    \subfloat{{\includegraphics[width=0.45\linewidth]{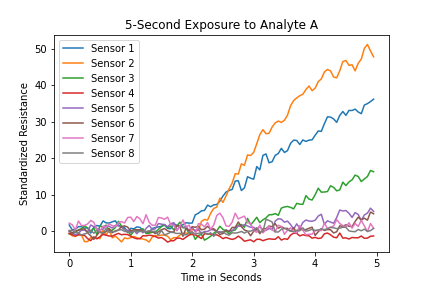} }}%
    \qquad
    \subfloat{{\includegraphics[width=0.45\linewidth]{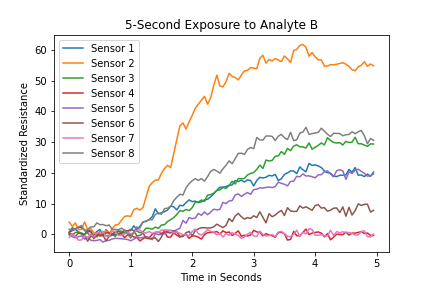} }}%
    \caption{Two five second single-analyte exposures to different analytes at the same concentration. Discrepancies in sensor resistance are explained by adsorption interactions between analytes and sensor coatings.}%
    \label{fig:two_singles}%
\end{figure*}



\subsection{ChemTime}\label{sec:intro_chemtime}

\textit{ChemVise} \cite{moore2023chemvise} introduced a molecular-semantic latent space styled on combined natural image-natural language spaces as in DeVise \cite{frome2013} and DALL-E \cite{dalle1} for improved classification of out-of-distribution chemical analytes. Though \textit{ChemVise} demonstrated how a chemistry-informed latent space using a pretrained molecular target embedding model improved classification outcomes over baselines, it failed to leverage inherent time series structure in the data.

\textit{ChemTime} modifies the \textit{ChemVise} approach by replacing the tabular-data embedding model with an iterative time series embedder with a moving-target approach to signal embedding. At each time step, the iterative encoder model uses a recurrent neural network backbone to encode the resistance signals of the input. At each iteration, a linear projection layer maps the state vector to a point in the molecular embedding space. The model loss is given by the sum of losses over the sequence, $\mathbb{L} = \sum \mathbb{L}_t$ where $\mathbb{L}_t$ may be given by an embedding distance such as a cosine distance, euclidean distance (MSE Loss), or bespoke representation loss as in DeViSE \cite{frome2013}.

\textit{ChemTime} has multiple benefits over the predecessor architecture. Utilizing sequences of representations in a meaningful latent space yields improved classification outcomes, inference during testing, and earlier classification compared to the fixed-window approach. Results sections will demonstrate the benefits of a light-weight approach for rapid classification, and the benefits of inference and analysis when sequences of representations may be used for meta-classification in the chemistry-informed latent.

\begin{lstlisting}[language=Python]
# Require:
# chemception_enc: Chemception with final linear layer removed
# RNN: One-layer iterative neural network
# lm: Linear projection from RNN state to target representation dimension
# chem_start: time index where analyte exposure begins
# criterion: Euclidian distance, cosine distance, DEViSE distance

for signal, label in train_loader: # load a batch of labelled sensor samples
    # Encode training labels to analyte-space. Before analyte flux begins, label should be an innate gas
    y_rep_seq = [chemception_enc(label) if i > chem_start else chemception_enc('Nitrogen') for i in signal.sequence_length]
    
    # Iterate over sequence with RNN
    x_seq = RNN(x)

    # Project RNN state sequence to representation space
    y_hat_seq = [lm(x) for x in x_seq]
    
    # Classification metric to target embeddings
    loss = criterion(y_hat_seq, y_rep_seq)

# Downstream classifier on embedded signals
# Use final output of projection for boosted input
emb_xtrain = lm(RNN((x_train))[-1]
opt_model = SVC.train(emb_xtrain, ytrain)

# Use the classifier to predict final representation of testing signals
emb_xtest = lm(RNN(x_test))[-1]
emb_ytest_hats = opt_model.predict(emb_xtest)
test_score = classification_metric(emb_ytest_hats, ytest)
\end{lstlisting}

\subsection{Implementation Details}

\subsubsection*{Label Sequence Generation} \textit{ChemTime} requires rephrasing a chemical exposure label into a sequence of targets in a chemistry representation space. Data are provided with labels corresponding to the concentration of the analyte exposure. For example, an exposure of 17\% Analyte B will be labelled [0,17,0,0]. We build a sequence of targets by first getting the representation of Analyte B as the activations of Analyte B in a chemistry representation model. For the purposes of \textit{ChemTime}, we utilize \textit{Chemception}, though any molecular representation could be used \cite{goh2017}. This representation is then unrolled into a sequence by concatenating a 'None' representation (the activations of the inert gas Nitrogen under the same molecular representation model) for each time step in which the flux of analyte vapor has not begun, and the representation at all subsequent time steps.

\subsubsection*{RNN and Linear Projection} \textit{ChemTime} uses a two-stage projection to iterate over a resistance signal and map to the chemistry-informed latent space.

\subsubsection*{Boosting}Subsequent to training the \textit{ChemTime} sequence embedding model, a tabular machine learning model may be fit on the final representations in the sequence. This is to classify the samples, which can be done with a naive nearest-target approach, or with a learned decision boundary. \textit{ChemTime} results discussed here use a simple SVC with a binary decision discriminating Analyte A samples from samples containing Analytes B, C, and D.

\subsection{Time Series Classification}\label{sec:time_series_classification}
Time series data are sequences of features with some inherent ordering, such as stock closing prices for each day of a year. Time series classification is highly relevant in applications across all domains in the physical sciences, manufacturing industries, technology, and more \cite{madan2018predicting, lin2005approximations, jonsson2004timesat, li2011throughput, semenick2000time, gupta2020approaches}. From sensors \cite{weiss2018applications}, video analysis \cite{jonsson2004timesat}, computer network traffic analysis \cite{madan2018predicting}, biomedical monitoring \cite{lin2005approximations}, manufacturing \cite{li2011throughput}, and airline industry efficiency \cite{semenick2000time}. The structural ordering of the features affects the discriminability of samples in the same manner as neighboring pixels in a natural image determine meaning to a human viewer \cite{bagnall2017great}. This ordering does not need to be through time: in the physical sciences, features may be ordered by frequencies, magnitudes, any ordered axis.

One univariate time series observation \textit{s} is a sequence of ordered pairs $(timestamp, value)$ \cite{xing2012early}. Multivariate time series (MTS) extends time series data to contain multiple features at each timestep of a signal, where the time series is a list of vectors over $d$ dimensions and $n$ observations. Data set $\mathbb{X}$ is given by observations $<X_1, ..., X_n>$. The $t$-th time index of the $i$-th sample of dimension $k$ is the scalar $x_{i,t,k}$ \cite{ruiz2021great}. Rapid classification of multivariate time series will call upon subsequences of $\vec{x}$. The \textit{length-l} prefix subsequences of sample $X_i$ are $X_i[:,1,t]$, or the first $t$ time observations of sample $X_i$ for all features. Subsequences are sometimes called \textit{shapelets} in the literature and are used as discriminative sub-elements of signals with distance metrics \cite{bostrom2017binary}.

\section{Study of MTSC Classifiers}\label{chpt:bakeoff}
We investigate a broad array of approaches to multivariate chemiresistive time series classification including at least one representative from all popular algorithm types to generate an up-to-date and competitive pool of models \cite{ruiz2021great, ottervanger2021multietsc}. We propose a variety of experiments in which the inductive biases and classification capabilities of each model are compared over a large swathe of real-world chemical sensing data sets, with additional analyses that compare model performance for challenges unique to the chemical sensing domain. Section \ref{sec:bo_results} compares the supervised classification performance of the classifiers across a variety of chemical sensing data sets in order to compare typical model performance by factoring out data biases. Section \ref{sec:rapid_classif} and \ref{sec:ttt} emphasize the importance of the rapid classification at testing time for chemical analyte detection, which is vital for a variety of health and safety devices. Relative model performances are weighted with their time to train and infer in Section \ref{sec:ttt}. Additional results in Section \ref{sec:inference} investigate confidence and discriminability of out-of-distribution analytes as a benefit of \textit{ChemTime}.

\subsection{Included Models}\label{sec:bakeoff_models}
\begin{table*}[t]
  \centering
  \begin{tabular}{|l|c|c|r|}
\toprule
Model & Type & Application & Source \\
\midrule

FullyConvolutionalNetwork & Conv. Deep Learning & Multivariate & \cite{zhao2017convolutional}\\
ConvolutionalNeuralNetwork & Conv. Deep Learning & Multivariate & \cite{zhao2017convolutional}\\
ROCKET & Random kernel transform & Multivariate & \cite{dempster2020ROCKET} \\
Catch22 & Feature extraction transform & Multivariate & \cite{lubba2019catch22}\\
RandomInterval & Concatenates random intervals & Multivariate & \cite{loning2019sktime}\\
CanonicalIntervalForest & Catch22 on random intervals & Multivariate & \cite{middlehurst2020canonical}\\
HIVECOTEV2 & Hierarchical transform ensemble & Multivariate & \cite{middlehurst2021hive}\\
ShapeletTransform & Rotation Forest on shapelet transform & Multivariate & \cite{bostrom2017binary}\\
WEASELMUSE & Bag-of-patterns for intervals & Multivariate & \cite{schafer2017multivariate}\\

BOSSEnsemble & Bag of Symbolic Fourier Symbols & Univariate & \cite{schafer2015boss}\\
MatrixProfile & Unified motifs and shapelets & Univariate & \cite{yeh2018time}\\
ContinuousIntervalTree & Information based decision tree & Univariate & \cite{deng2013time}\\
RotationForest & Forest on random PCA transforms & Univariate & \cite{rodriguez2006rotation}\\

MultiLayerPerceptron & FCN Deep Learning & Non-temporal & \cite{wang2017time} \\
SVC & Support vector classifier & Non-temporal & \cite{cortes1995support}\\
KNeighbors & K-neighbors supervised clustering & Non-temporal & \cite{fix1989discriminatory}\\
GaussianProcess & Gaussian process classification & Non-temporal & \cite{seeger2004gaussian}\\
DecisionTree & Information-splitting tree & Non-temporal & \cite{quinlan1986induction}\\
RandomForest & Forest of decision trees & Non-temporal & \cite{ho1995random}\\
AdaBoost & Boosting ensemble & Non-temporal & \cite{freund1997decision}\\
GaussianNB & Gaussian Naive Bayes classifier & Non-temporal & \cite{chan1982updating}\\
QuadraticDiscriminantAnalysis & Quadratic classifier by Bayes' rule & Non-temporal & \cite{rao1948large}\\
\bottomrule
  \end{tabular}
  \caption{Competitive classifiers for chemiresistive sensor array classification, a brief description, and their origin.}
  \label{tab:bakeoff_models_source_table}
\end{table*}

A diverse cast of models representing multiple approaches to univariate and multivariate time series classification are drawn for our chemical sensing survey. Each is given a brief description in Table \ref{tab:bakeoff_models_source_table}. Univariate time series models are adapted to the multivariate time series classification paradigm with each of two algorithms detailed in Section \ref{sec:univariate-multivariate}. Non-temporal machine learning models are adapted using a tabularization of the multivariate time series data with column concatenation described in Section \ref{sec:univariate-multivariate}. We defer to \cite{ruiz2021great} for algorithm details of contemporary classifiers, and their respective publications for exact implementation.

\subsection{Classification Study}\label{sec:bo_results}
In order to provide optimal model classes to chemiresistive sensor researchers for multiple analyte discrimination, we must evaluate the effectiveness of multivariate time series classifiers for general chemiresistive sensor hardware. We estimate classifier performance for a general chemiresistive sensor array by studying eleven different sensor configurations, each with unique surface chemistries. Model performance on these eleven distinct data sets demonstrates the efficacy of each classifier family for the broader domain of chemiresistive sensor array classification.

We study the supervised performance of classifier families with the following process. Each classifier is trained four times on each of eleven real-world chemiresistive sensor array data sets. Each of these four training iterations uses $75\%$ split on the training data where a 25\% fold is removed from the training corpus for that classifier split. Then each of these split models is validated on a holdout set of testing data corresponding to that experiment which does not include training split samples nor withheld split samples. This is performed for each model described in Table \ref{tab:bakeoff_models_source_table}, \textit{ChemTime} and 11 chemical sensing data sets described in Section \ref{sec:chemical_sensing}.

As in the earlier empirical survey \textit{The Great Multivariate Time Series Classification Bake Off} \cite{ruiz2021great}, each model uses a parameterization taken from K-Fold hyperparameterization on a domain-diverse set of classification tasks as provided in the {\fontfamily{qcr}\selectfont\ sktime} repository \cite{loning2019sktime}. An empirical study of multivariate time series classifiers must consider the balance of computation, time to train and infer, and ultimate model performance. For the sake of this study, each model is trained and validated once under default hyperparameterization. 
\begin{figure*}
    \centering
    \includegraphics[width=0.7\linewidth]{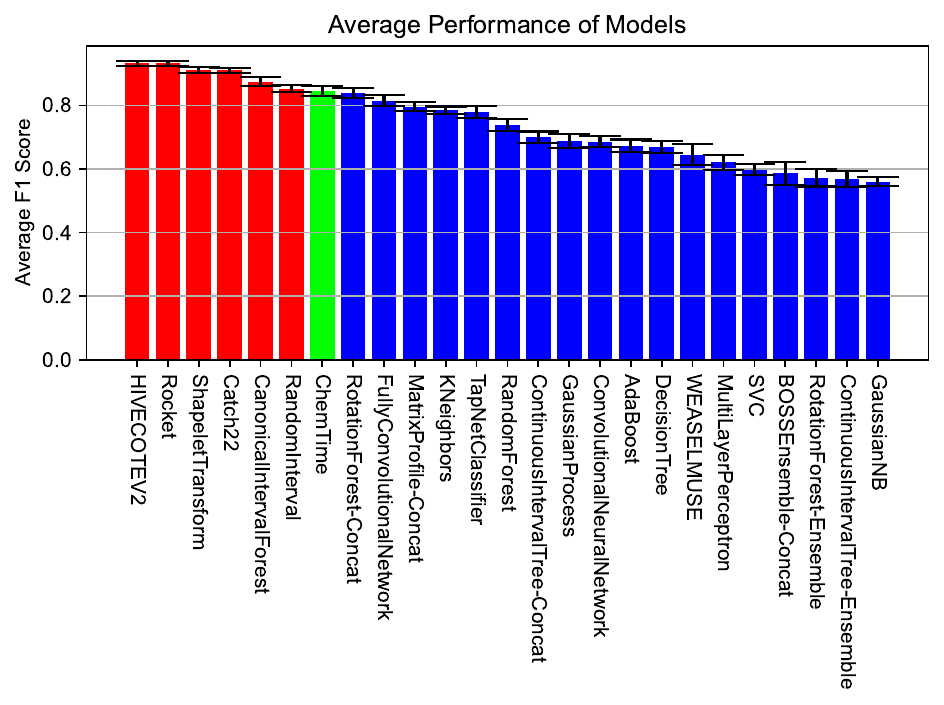}
    \caption{Average model performances across 44 splits of 11 chemiresistive sensor array data sets.}
    \label{fig:avg_model_scores}
\end{figure*}

\begin{figure*}
\begin{center}
\includegraphics[width=0.7\linewidth]{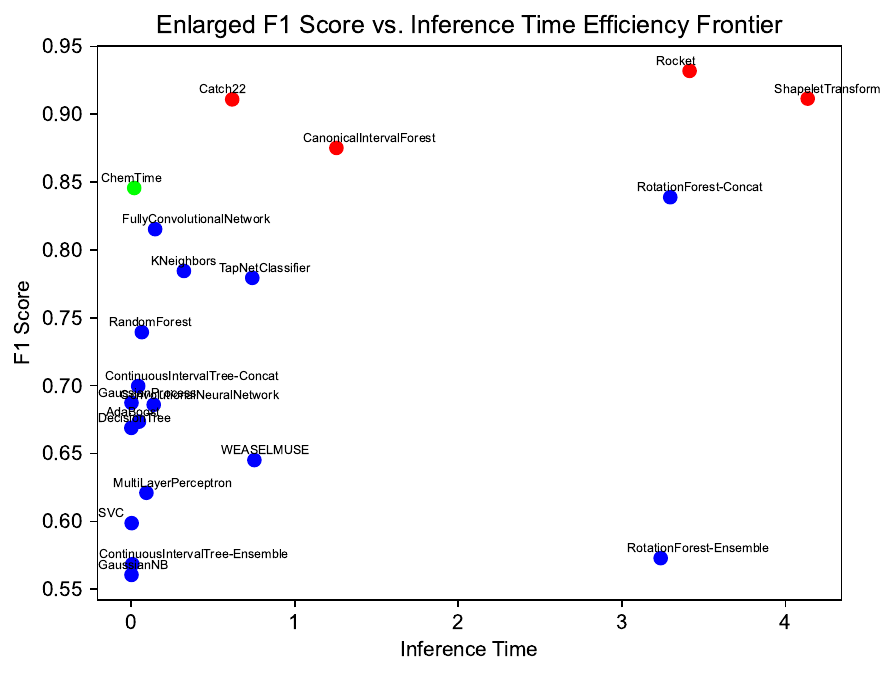}
\caption{Classification performance frontier versus time of inference. Longer inference times may lead to improved classification outcomes at the cost of a delayed classification decision.} \label{fig:tte_vs_score}
\end{center}
\end{figure*}

\begin{figure*}
    \centering
    \includegraphics[width=0.7\linewidth]{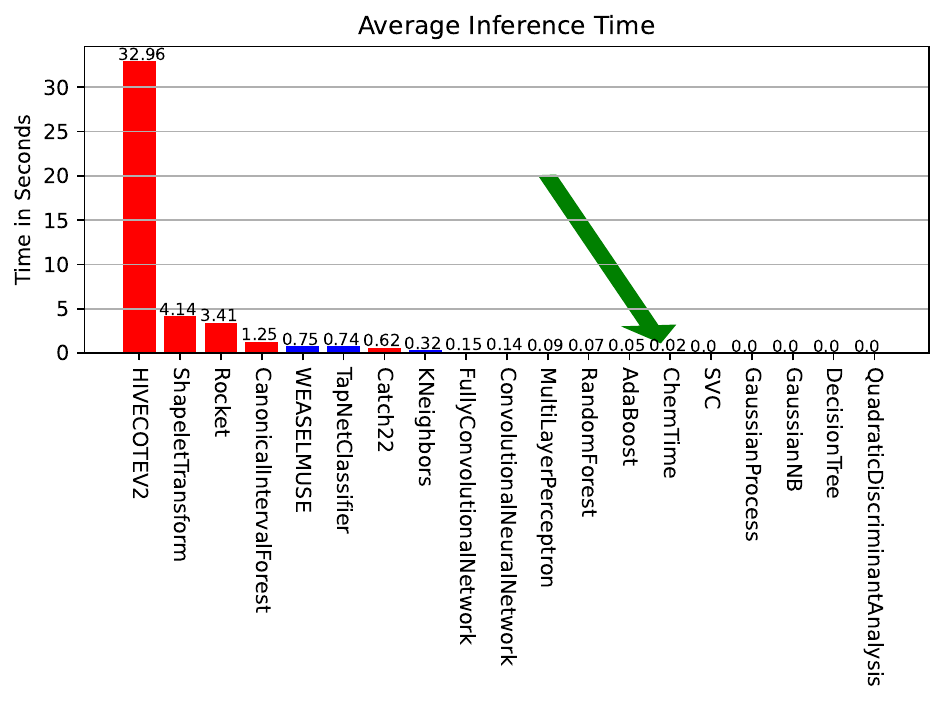}
    \caption{Wall-clock time in seconds to test each model on a testing set of 32 holdout trials.}
    \label{fig:ttt}
\end{figure*}

Figure \ref{fig:avg_model_scores} visualizes the mean performance of models across all eleven experimental data sets, with four splits of each training data set. Some models fail to converge for many data sets and have been removed from subsequent analyses (Matrix Profile with Column Ensembles, BOSS Ensemble with Column Ensembles, and Quadratic Discriminant Analysis). Our hypotheses on the chemiresistive data structure have previously assumed shapelets or time warping KNN would be the most successful models. Though these are effective, random transformations with linear classifiers (ROCKET) as well as complex feature ensembling (HIVECOTE) outperform a variety of shapelet-based approaches on average.

\begin{figure*}
\includegraphics[width=0.9\textwidth]{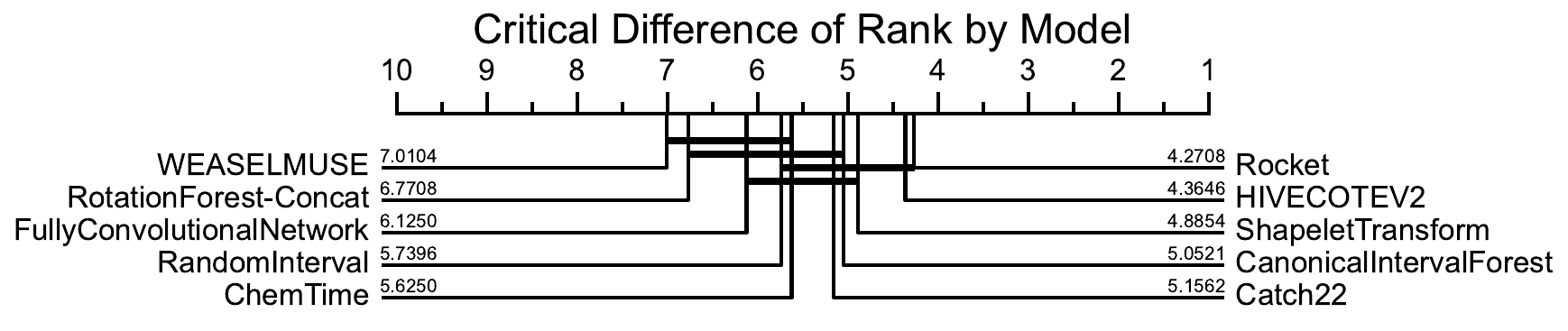}
\caption{\label{fig:mcc_cd}Critical difference plot demonstrating cliques of top 10 models by rank distribution. Clique bars indicate statistical uncertainty of difference in average rank of clique members based on performance across sensor data sets.}
\end{figure*}

Figure \ref{fig:mcc_cd} shows the critical difference between the top 10 model average ranks over 44 splits of the 11 sensor array training sets. The horizontal bands link a clique of models with statistically insignificant difference of average rank given performance across all splits. In the top-performing clique we find models from each of the primary families of time series classifiers - random kernels, feature ensembles, shapelets, decision trees, and deep learning \cite{gupta2020approaches}.

\subsection{Rapid Classification}\label{sec:rapid_classif}
\begin{figure}
    \centering
    \includegraphics[width=\linewidth]{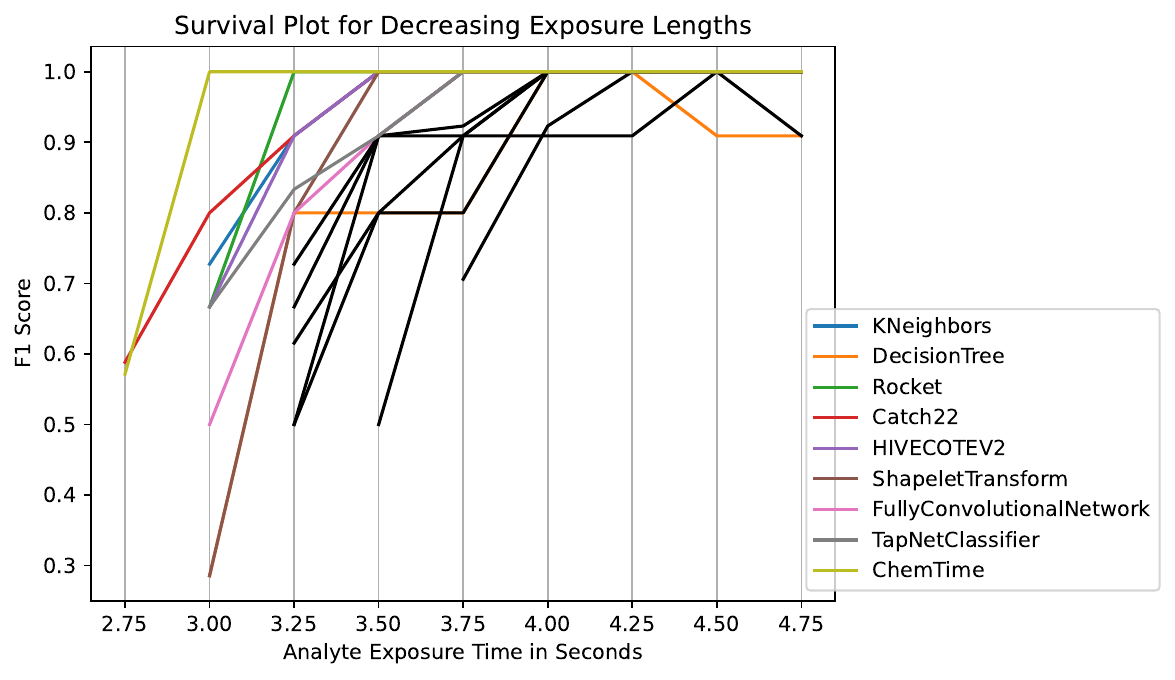}
    \caption{F1 scores of top-10 models trained on incrementally decreasing exposure times. Models which fall below an F1 score of 0.8 on the testing set for an exposure time are eliminated.}
    \label{fig:survival_plot}
\end{figure}

\begin{figure}
    \centering
    \includegraphics[width=\linewidth]{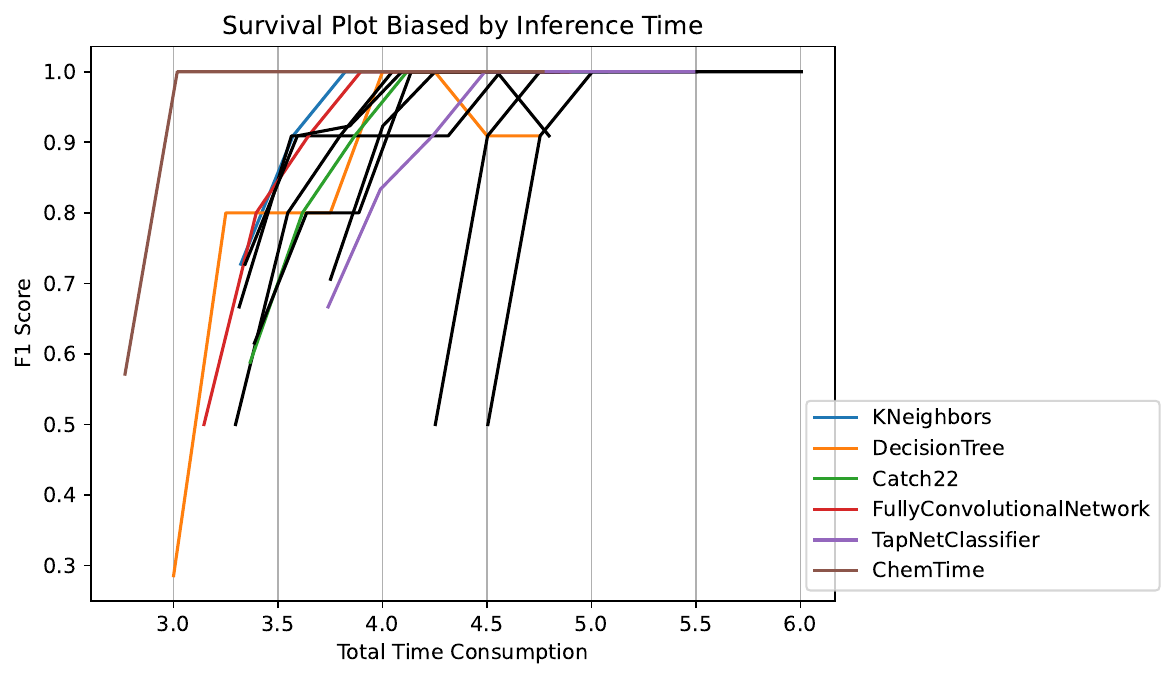}
    \caption{Survival plot biased by the inference times factored into the length of exposure. When accounting for the processing time of the model as part of the length of exposure, the improvement of \textit{ChemTime} is remarkable. Furthermore, \textit{ChemTime} benefits from real-time signal processing which many others do not.}
    \label{fig:survival_bias_plot}
\end{figure}

Research in the rapid detection of harmful chemical analytes often incentives rapid classification for improved safety devices. Here we propose a contest in the rapid classification of chemical Analyte A in increasingly challenging foreshortened signals. This contest begins by comparing classifier testing set scores for 5-second resistance curve exposures. For each iteration of the test, models with an average F1 score below 0.8 are eliminated, and the length of the exposure window is reduced by 0.25 seconds. The experiment is then repeated until no models remain. Figure \ref{fig:survival_plot} visualizes the results of our experiment in the survival of models in retaining high classification accuracy during increasingly challenging rounds of chemical sensing by decreased exposure windows.

An early classifier $\textbf{C}$ is \textit{serial} if $\textbf{C}[s[1,l_0]) = \textbf{C}(s[1,l_0+i])$ for any $i>0$ - that is, the classifier $\textbf{C}$ does not benefit from longer prefixes of the data and will not change the estimate with further steps \cite{xing2012early}. Investigating Figure \ref{fig:survival_plot} in reverse, we find the best classifiers to be serial given exposure times of around 3.25 to 3.5 seconds, and do not benefit from further exposure to signals. Classifiers which remain serial to very brief exposure windows include ROCKET, Catch22, HIVECOTEV2, and RandomInverval classifiers. These represent a surprisingly diverse mix of shapelet-based approaches, model ensembles, and transformation classifiers, and demonstrates that shapelet-based approaches though inductively sound for the data may struggle with variance in sensor responses and generalizability.

\subsection{Training and Inference Time}\label{sec:ttt}

To account for discrepancies in training and validation time, we quantify the trade-off between training and performance in order to account for the benefits of hyperparameter tuning for the full empirical study in Figure \ref{fig:full_frontier}.


Figure \ref{fig:ttt} compares the time to test each model. Outlaying models include the MatrixProfile-Concat, HIVECOTEV2, and FullyConvolutionalNetwork. The MatrixProfile-Concat training time demonstrates one shortcoming of the column concatenation technique of univariate classifier adaptation which is the significant increase in compute for dimensionality compared to the MatrixProfile-Ensemble. In addition the FullyConvolutionalNetwork would train significantly faster on a graphical processing unit, but specialty hardware may not be available for a real-world deployed tool with limited processing and power requirements. 


An inference time plot also tells a very important story. When we deploy a hardware chemical sensing device, the amount of compute may be very limited. The time to process the inference of a sample would need to be factored into the rapid classification scores as the processing of the testing sample would delay reporting the classification result, resulting in wasted time in a safety situation.

To weigh the importance of time to train and test against accurate predictions, we visualize the spread of model testing time and testing performance in Figure \ref{fig:tte_vs_score}.There exist extreme outliers in the time to train and test (Appendix full plot) with significant competition from models with slightly lower accuracy but orders of magnitude faster time to train.

Figure \ref{fig:tte_vs_score} investigates the trade-off between inference time and testing set performance. As indicated in Section \ref{sec:rapid_classif}, the rapid detection of chemical analytes is vital to multiple applications. By investigating the distribution of scores against the model inference times we identify the existence of a frontier in the trade off between classification time and performance. Some models lay along this frontier, but some are inferior in both metrics. Depending on the urgency of the classification, any model along the frontier may be selected when a combination of rapid classification and accuracy are appropriate for the sensing task. In real-world chemical sensing edge devices, rapid classification may be vital.


\subsection{Semantic Inference}\label{sec:inference}
\textit{ChemTime} has the unique benefit of exploiting domain knowledge of molecular semantic properties. Translating from chemical resistance space to a meaningful semantic space implies improved inferential possibilities including early and rapid classification as well as out-of-distribution classification. For this reason we investigate possible advantages of the \textit{ChemTime} approach for chemical sensing.

\begin{figure*}
    \centering
    \subfloat{{\includegraphics[width=0.45\linewidth]{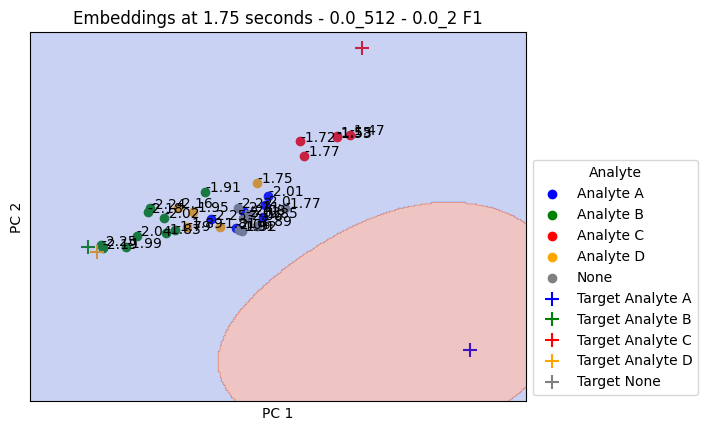} }}%
    \qquad
    \subfloat{{\includegraphics[width=0.45\linewidth]{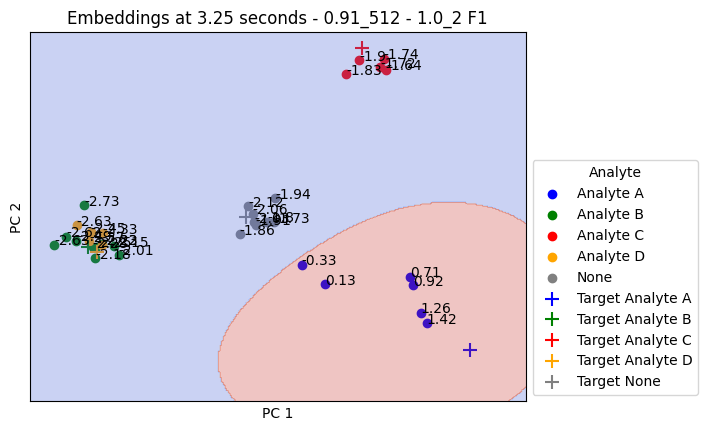} }}%
    \caption{Iterating through time updates the positions of sample embeddings at every time step. Animated visual available online.}%
    \label{fig:chemtime_moving}%
\end{figure*}

Figure \ref{fig:chemtime_moving} demonstrates how \textit{ChemTime} iterates over a signal to update the molecular representation in the chemistry latent space. A simple learned projection maps from the internal state of the RNN to the chemistry latent and updates at each resistance timestep. The relative distances between sample labels in the chemistry-informed latent yields further inference on the sequence of representations at testing time.

Figure \ref{fig:decision_dists} demonstrates how a sequence of representations at inference time yields additional benefits over a typical black-box classifier. Early prediction and confidence inference are implicit to the design of \textit{ChemTime} in which the boosting classifier yields a distance from the decision boundary. While this distance is inherent to simple classifiers such as Support Vector Machines, the two-stage classifier of a learned representation alongside the secondary classifier allows inference as well as a high performance, evidenced by the results in Section \ref{sec:bo_results}.

\begin{figure}
    \centering
    \includegraphics[width=\linewidth]{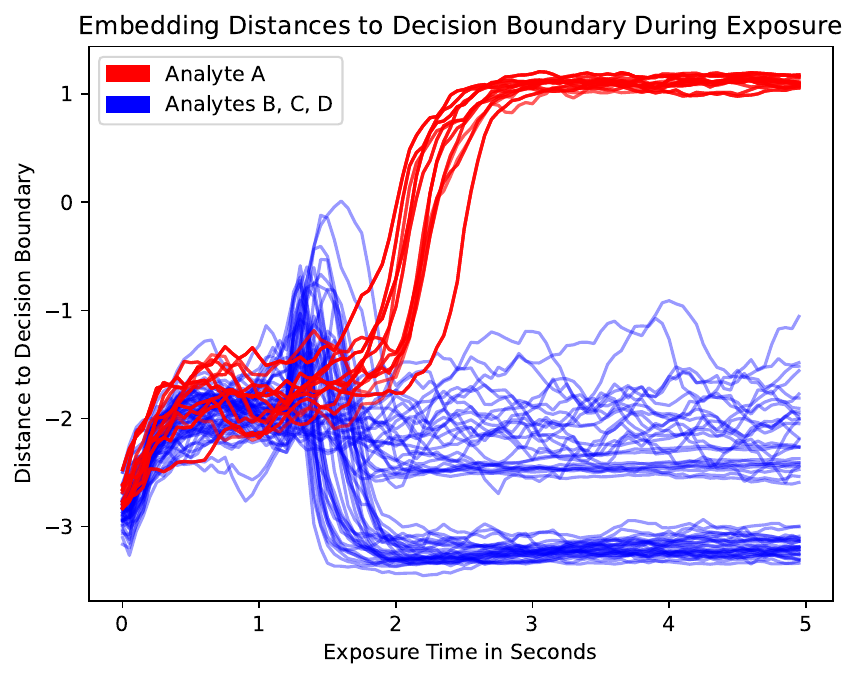}
    \caption{Distances through time from boosting classifier decision boundaries give inference over samples. A validation set may determine the optimal early classification window for testing set samples based on embedding trajectory.}
    \label{fig:decision_dists}
\end{figure}

\subsection{Discussion}

Real-world utility of machine learning for the physical sciences extends past chemiresistive sensor arrays to countless applications. Addressing the inductive biases of a variety of models is particularly motivating when we also have access to a variety of experimental protocols and sensors, which themselves make assumptions about the nature of discrimination. By replicating the experimental design of Ruiz \cite{ruiz2021great}, we find that as a trend models which perform well on the diverse UEA archive \cite{UEA2018} of multivariate time series classification data sets also perform well on our chemical sensing task. Empirical survey benchmark results fail to find a clear pattern for successful classifier algorithms which differentiate the multivariate sensor array task from a generic multivariate time series classification task. Specialized hardware for the detection of particular chemical analytes for specific industry or safety applications may alter this pattern to highlight the effectiveness of alternative methods.

We find that a \textit{ChemTime} model with biases designed for the task improves the efficiency for the inference-accuracy trade off identified in Figure \ref{fig:full_frontier}. The inductive biases of the \textit{ChemTime} architecture and loss combined with the inference of a boosting classifier yield a performant and extremely rapid classifier with unique advantages against the field in terms of adaptability and flexibility to a field of additional techniques in inference and analysis of representation sequences.

Furthermore, we find an efficient frontier balancing the time it takes for models to perform inference against the supervised learning performance of those classifiers. We determine a similar pattern exists in the training and hyperparameter optimization times for these same classifiers. Finally, we discuss extensions of classifiers to the rapid analyte classification task, and outline how many of the contemporary approaches to time series classification are inappropriate for rapid classification or incompatible with early classification, a relevant subdomain of sensor classification.

\section{Acknowledgements}
This manuscript has been authored with funding provided by the Defense Threat Reduction Agency (DTRA). The publisher acknowledges that the US Government retains a nonexclusive, paid-up, irrevocable, worldwide license to publish or reproduce the published form of this manuscript, or allow others to do so, for US government purposes.

Approved for public release. Distribution unlimited.

\appendix

\section{Appendix}
Appendix one here. Additional results including training time for hyperparameter sweeps, univariate vs. multivariate conversion. Additional images can go here - analysis and conclusions leave in text.

\begin{figure}
    \centering
    \includegraphics[width=\linewidth]{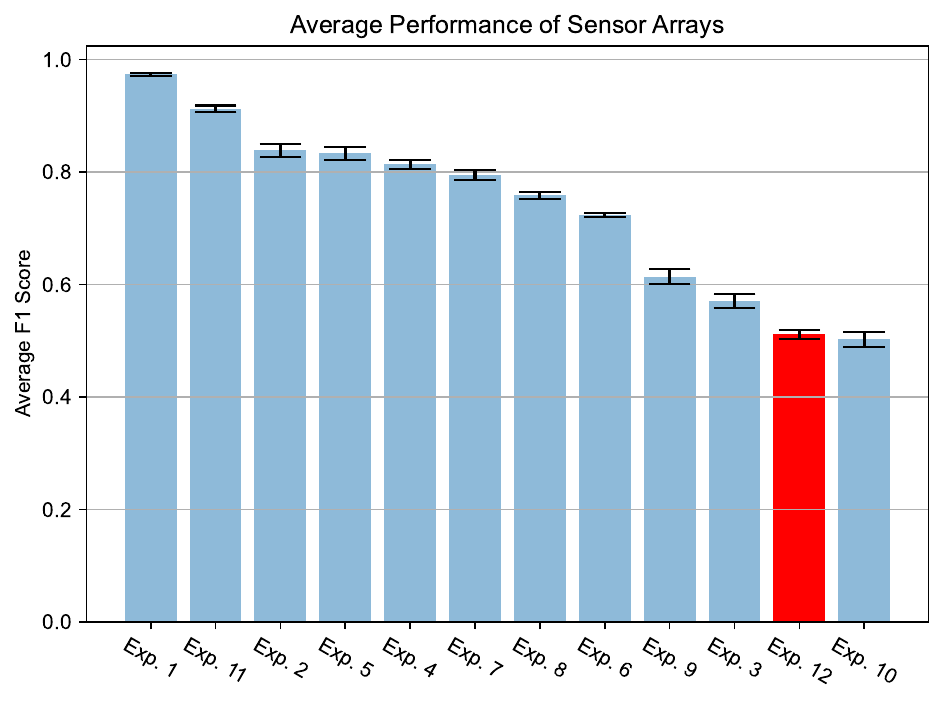}
    \caption{Average sensor array performance given by the mean of all model performances over four training splits. Outlying models removed.}
    \label{fig:avg_exp_scores}
\end{figure}

In addition to investigating the average performance of models across a variety of domain datasets, we are in a unique position in developing an effective chemical sensing tool at the hardware level. In addition to optimization of model selection across datasets, we seek an effective \textit{dataset} which leads to successful analyte detection models at testing time. One approach to finding an optimal sensor array set is to compare the average performance of diverse models learning outcomes on the dataset (Figure \ref{fig:avg_exp_scores}). The corresponding testing accuracy of a model trained on a particular set gives a sense of the predictability and discriminability of the sensors, while ``factoring out`` the inductive biases of each model by considering a diverse set of classifiers.

\begin{figure*}
    \centering
    \includegraphics[width=\linewidth]{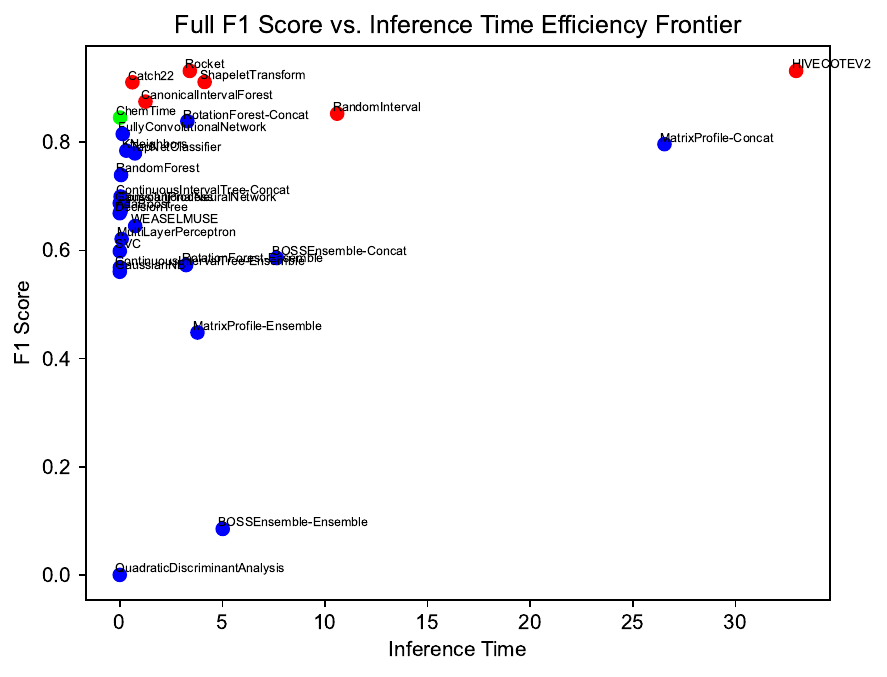}
    \caption{Full time to infer frontier.}
    \label{fig:full_frontier}
\end{figure*}

\subsection{Univariate Adaptation Algorithms}\label{sec:univariate-multivariate}

\begin{figure}
    \centering
    \includegraphics[width=\linewidth]{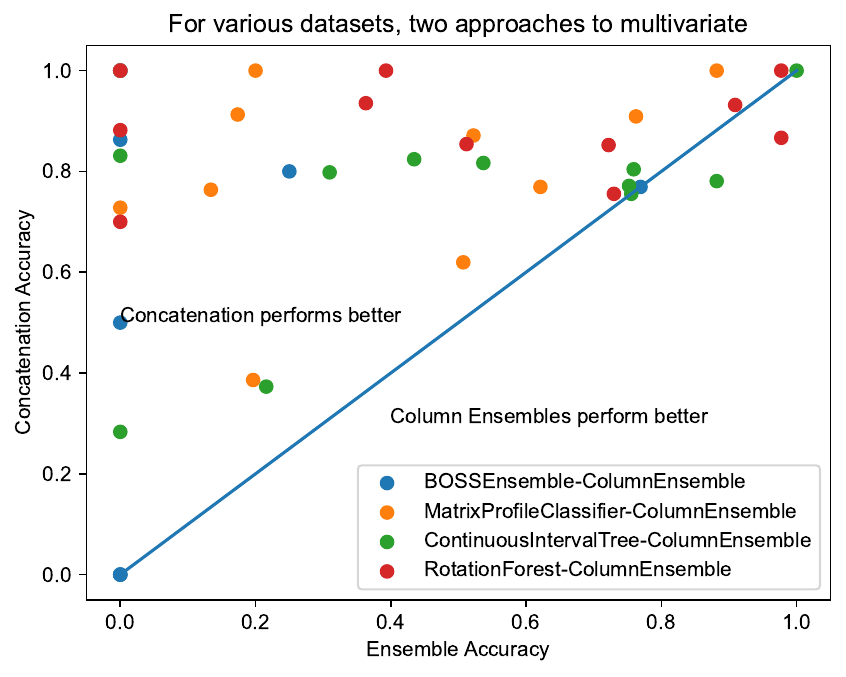}
    \caption{Two approaches to multivariate adaptation of univariate time series classifiers. Above the line, the column concatenation approach outperforms the column ensembling approach for the dataset and model.}
    \label{fig:two_approaches}
\end{figure}

Univariate time series classification models are a well-researched literature and may be adapted to multivariate time series classification tasks with two alterations to the training algorithm of arbitrary univariate classifiers. These adaptations include the following:

\begin{enumerate}
    \item Column Concatenation: given training data array $\mathbb{D}$ of $n$ samples of $k$ dimensions of length $t$ $(n, k, t)$ and classifier $C$, reshape the data by concatenating dimension $d_i$ to the last element of dimension $d_{i-1}$, yielding array of shape $(n, k*t)$:

    \begin{align}
    &(n,k,t) \rightarrow \\
    &(n_{1,1}, ..., n_{1,t}, ..., n_{2,t}, ... n_{k,1}, ..., n_{k,t}) 
    \end{align}
    
    Then train classifier $C$ on resulting array $\mathbb{D}^*$ of shape $(n, k*t)$
    
    \item Column Ensembling: given training data array $D$ of $n$ samples of $k$ dimensions of length $t$ $(n, k, t)$ and classifier $C$, define $k$ training subarrays $(n,d_1,t), (n, d_k, t)$ for each dimension $d$ of $k$.
    
    Train classifier $C_i$ on each resulting array $\mathbb{D}_i = (n, d_i, k)$, and use ensemble voting to classify the sample \cite{polikar2006ensemble}.
\end{enumerate}

These approaches have the advantage of relying on a rich literature of univariate time series classification, but each has a substantial downside. Column concatenation (1) substantially increase the dimension of the feature space, which is particularly disadvantageous for decomposition classifiers such as matrix profile classifiers as well as increasing the effect of the curse of dimensionality for high dimensional spaces. In addition this removes the option to early-classify samples \cite{xing2012early, hartvigsen2019adaptive} and is incompatibility with state-based approaches such as an RNN. Second, the column ensembling approach (2) may train multiple classifiers which each fail to accurately predict the totality of the sample given one channel of information, as we assume with our chemical sensing data in which the \textit{diversity} in sensor coating affinities yields discriminability. Individual sensors do not contain enough information for a classifier to make a reasonable prediction.



\bibliographystyle{IEEEtran}
\bibliography{generalizing_sensors_bib.bib}

\end{document}